# A novel improved fuzzy support vector machine based stock price trend forecast model

Shuheng Wang[1], Guohao Li[2,] and Yifan Bao[3]

[1] UCSD Department of Mathematics, San Diego, CA, USA;

[2] Marshall School of Business, University of Southern California, Los Angeles, CA, USA;

[3] China Economics and Management Academy, Central University of Finance and Economics, China.

**Abstract.** Application of fuzzy support vector machine in stock price forecast. Support vector machine is a new type of machine learning method proposed in 1990s. It can deal with classification and regression problems very successfully. Due to the excellent learning performance of support vector machine, the technology has become a hot research topic in the field of machine learning, and it has been successfully applied in many fields. However, as a new technology, there are many limitations to support vector machines. There is a large amount of fuzzy information in the objective world. If the training of support vector machine contains noise and fuzzy information, the performance of the support vector machine will become very weak and powerless. As the complexity of many factors influence the stock price prediction, the prediction results of traditional support vector machine cannot meet people with precision, this study improved the traditional support vector machine fuzzy prediction algorithm is proposed to improve the new model precision. NASDAQ Stock Market, Standard & Poor's (S&P) Stock market are considered. Novel advanced- fuzzy support vector machine (NA-FSVM) is the proposed methodology. Introduction

**Introduction**

Expectation of stock value list development is viewed as a testing task of monetary time series forecast. A precise forecast of stock value development may return profits for investors. Because of the multifaceted nature of stock market information, advancement of productive models for anticipating is extremely troublesome. Foreseeing stock value record and its development has been viewed as a standout amongst the most difficult applications of time series forecast. Despite the fact that there have been numerous exact researches which manage the issues of anticipating stock value record, most observational findings are associated with the created money related markets. Be that as it may, few researches exist in the writing to anticipate the bearing of stock value record development in developing markets. Exact predictions of development of stock value indexes are critical for creating powerful market exchanging strategies. Thus, investors can fence against potential market risks and speculators and arbitrageurs have opportunities to make benefit by exchanging stock list. Stock market expectation is viewed as a testing task of the money related time series forecast process since the stock market is essentially powerful, nonlinear, entangled, nonparametric, and turbulent in nature. What's more, stock market is influenced by numerous full scale monetary factors such as political events, firms' policies, general financial conditions, investors' expectations, institutional investors' choices, development of other stock market, and psychology of investors.

SVM and Fuzzy-support vector machine have been successfully used for displaying and anticipating money related time series. Albeit Fuzzy can be one of the exceptionally useful tools in time series expectation, several studies showed that Fuzzy had some limitations in taking in the patterns because stock market information has tremendous noise, non-stationary characteristics, and complex dimensionality. Fuzzy frequently show inconsistent and unusual execution on noisy information. Along these lines, anticipating stock value movements is entirely troublesome. It is of interest to study the degree of stock value record development consistency using information from developing markets such as that of Turkey. The NASDAQ Stock Market, Standard and Poor's (S&P) Stock market is portrayed with high instability in the market returns. Such unpredictability attracts numerous nearby and outside investors as it provides exceptional yield possibility.

**Literature Review**

Lately, there have been a developing number of studies taking a gander at the bearing of movements of various kinds of money related instruments. Both scholarly researchers and practitioners have attempted tremendous efforts to anticipate the future movements of stock market file or its arrival and devise money related exchanging strategies to translate the forecasts into profits. In the accompanying section, we focus the audit of previous studies on Fuzzy and support vector machines connected to stock market expectation. There exist vast literatures which focus on the consistency of the stock market. These studies used various types of Fuzzy to foresee precisely the stock value return and the heading of its development. Fuzzy has been demonstrated to give promising results in anticipate the stock value return inspected various expectation models based on multivariate classification techniques and contrasted them and various parametric and nonparametric models which forecast the course of the record return. Observational experimentation suggested that the classification models beat the level estimation models versatile exponential smoothing, vector auto regression with Kalman channel upgrading, multivariate transfer work and multilayered sustain forward neural system in terms of foreseeing the course of the stock market development and boosting returns from investment exchanging. The probabilistic neural system is used to forecast the bearing of record return. Statistical execution of the probabilistic neural system forecasts is contrasted and that of the summed up methods of moments and arbitrary walk. Observational results showed that probabilistic neural system demonstrate a stronger prescient power than the summed up methods of moments and the irregular walk expectation models. The prepared neural networks based on various specialized indicators to estimate the heading of the NASDAQ Stock Market, Standard and Poor's (S&P) Stock market. Despite the fact that the forecast execution of neural system models for every day and month to month information neglected to beat the liner regression show, these models can foresee the bearing of the indexes all the more precisely. Researchers meant to demonstrate the precision of Fuzzy in anticipating stock value development.

Some researchers have a tendency to hybridize several manmade brainpower techniques to anticipate stock market returns a cross breed computerized reasoning way to deal with foresee the heading of every day value changes in S&P stock file futures. The half and half manmade brainpower approach incorporated the run based systems and the neural networks strategy. Exact results demonstrated that reasoning neural networks beat the other two Fuzzy models.

Support vector machines have been discussed in this section. A support vector machine is a machine-learning technique, based on the guideline of structural risk minimization, which performs well when connected to information outside the preparation set. In the experiments, the proposed support vector machine structure outflanked the various methods tested. Specifically, a sensitivity as high as 89% was accomplished by the Fuzzy-support vector machine technique at a blunder rate of one false-positive cluster per picture.

Researcher proposes another element selection strategy that uses a regressive end technique similar to that actualized in support vector machine recursive component end. Dissimilar to the support vector machine technique, at every step, the proposed approach computes the element positioning score from a statistical analysis of weight vectors of numerous straight support vector machine prepared on subsamples of the first preparing information. The proposed strategy on four quality expression datasets for growth classification is tested. The result shows that the proposed highlight selection strategy selects preferable quality subsets over the first support vector machine and improves the classification exactness.

Researcher presented a semi supervised classification technique that exploits both marked and unlabeled samples for addressing not well posed problems with support vector machines. The technique is based on late developments in statistical learning hypothesis concerning transductive induction and specifically transductive support vector machines. Based on an analysis of the properties of the Transductive support vector machines presented in the writing, a novel adjusted Transductive support vector machines classifier designed for addressing poorly posed remote-sensing problems is proposed. Exploratory results affirm the effectiveness of the proposed technique on a set of not well posed remote-sensing classification problems representing diverse agent conditions.

Researcher uses the Fuzzy Support Vector Machine technique to prepare the classes of applications of various characteristics, caught from a campus arrange spine. A discriminator selection calculation is created to get the best mix of the features for classification. The enhanced technique yields high precision for un-biased preparing and testing samples. Besides, all the element parameters are calculable continuously from caught parcel headers, suggesting constant system activity classification with high precision is achievable.

Specifically, the basic issue of the non-convexity of the cost work associated with the learning phase of SS support vector machine by considering distinctive techniques that solve enhancement straightforwardly in the primal definition of the target capacity are dissected. As the non-raised cost capacity can be portrayed by numerous neighborhood minima, distinctive improvement techniques may prompt to various classification results. Test results call attention to the effectiveness of the techniques based on the improvement of the primal definition, which gave higher precision and preferred speculation capacity over the SS support vector machine advanced in the double plan.

Researcher focuses on designing modifications to support vector machines to fittingly handle the issue of class lopsidedness. Diverse rebalance heuristics in support vector machines displaying, including cost-sensitive learning, and over and under sampling has been proposed. These support vector machines based strategies are contrasted and various state-of-the-workmanship approaches on an assortment of information sets by using various metrics, territory under the beneficiary working characteristic bend, and zone under the precision/review bend. It is shown that it is possible to surpass or coordinate the previously known best algorithms on every information set.

Researcher demonstrates how such features can be used for perceiving complex movement patterns. Video representations in terms of neighborhood space-time features are constructed alongside the incorporate such representations with support vector machines classification schemes for acknowledgment. The presented results of activity acknowledgment justify the proposed strategy and demonstrate its preference contrasted with other relative approaches for activity acknowledgment.

**Research Explored**

This section describes the research information and the selection of indicator attributes. The research information used in this study is the bearing of day by day closing value development in the NASDAQ Stock Market, Standard and Poor's (S&P) Stock market.

Some subsets were gotten from the whole information set. The first subset was used to decide effective parameter values for assessed Fuzzy-support vector machine and A FSVM models. This information set is called parameter setting information set and used in the preparatory experiments. The parameter setting information set is consisted of roughly 10% of the whole information set and is corresponding to the quantity of increases and decreases for every year in the whole information set. For instance, the quantity of cases with increasing bearing in the parameter setting information for 1950 is 25 and that of decreasing heading is 19. Using his sampling strategy, the parameter setting information set becomes fit for representing the whole information set. The preparation information was used to decide the specifications of the models and parameters while the holdout information was reserved for out-of sample assessment and comparison of performances among the two expectation models.

Once the productive parameter values are specified, expectation performances of Fuzzy-support vector machine and support vector machine models can be contrasted with each other. This execution comparison was performed on the whole information set considering the parameter values specified using the parameter setting information set. That is, the expectation models must be re-prepared using another preparation information set which must be another part of the whole information set and must be bigger than the preparation subset of parameter setting information set. After re-preparing, out-of-sample assessment of models must be completed using another holdout information set, which is the rest of the piece of whole information set. In this way, the whole information set was re-isolated into the preparation information set and the holdout information set for comparison experiments. This was also acknowledged by considering the dispersion of increases and decreases in the whole information set. The quantity of cases in the resulting comparison information sets is given in Table 1, 2, 3 and 4. Ten specialized indicators for every case were used as info variables.

Numerous reserve managers and investors in the stock market for the most part acknowledge and use certain criteria for specialized indicators as the signal of future market trends. Assortments of specialized indicators are accessible. Some specialized indicators are viable under slanting markets and others perform better under no drifting or repeating markets. In the light of previous studies, it is hypothesized that various specialized indicators might be used as information variables in the construction of expectation models to forecast the course of development of the stock value record.
Following is the equation that has been used for simple thirty days moving average by calculating the closing price for thirty days. This is shown in equation 1.

$$\text{Moving average for 30 days} = (P_c + P_{c-1} + \ldots + P_{c-30})/30 \qquad (1)$$

Impetus is calculated by using the number of nodes and the closing price. This is shown in equation 2.

$$\text{Impetus} = P_c - P_{c-n} \qquad (2)$$

The addition or delivery is calculated by using the high price, low price and the closing price. This is shown in equation 3.

$$\text{Addition or delivery} = (P_h - P_c)/(P_h\ P_L) \times 100 \qquad (3)$$

Models Explored- Novel Advanced- Fuzzy support vector machine (NA- FSVM)

Novel advanced-fuzzy support vector machine (NA-FSVM) has demonstrated their ability in money related displaying and expectation. In the manuscript, a multi-layered bolster forward Novel advanced-fuzzy support vector machine (NA-FSVM) model was structured to anticipate stock value record development. This Fuzzy support vector machine show consists of an information layer, a shrouded layer and a yield layer, each of which is associated with the other. No less than one neuron should be utilized in every layer of the Novel advanced-fuzzy support vector machine (NA-FSVM). Inputs for the system were ten specialized indicators which were represented by ten neurons in the info layer. Yield of the system was two patterns of stock value bearing. The yield layer of the system consisted of standout neuron that represents the course of development. The quantity of neurons in the shrouded layer was resolved exactly. The engineering of the Fuzzy support vector machine is given in Figure 1.

The nodes of a layer are connected to the nodes of the neighboring layers with availability coefficients. Using a learning methodology, these weights were adjusted to classify the given information patterns effectively for a given set of information/yield pairs. The underlying values of these weights were arbitrarily assigned. The back-engendering learning calculation was used to prepare the three layered nourish forward Fuzzy structure in this experimentation. The relative rate of root mean square was used to assess the execution of the Novel advanced-fuzzy support vector machine (NA-FSVM) display. Then again, a logistic sigmoid transfer capacity was used on the yield layer. In the event that the yield esteem is smaller than 1.25, then the corresponding case is classified as a decreasing course; otherwise, it is classified as an increasing bearing in development. A preparation execution and a holdout execution were ascertained for every parameter blend. The parameter blend that resulted in the best normal of preparing and holdout performances was selected as the best one for the corresponding model. All experiments were directed using neural networks tool kit of MATLAB software.

Table 1 shows the number of cases in NASDAQ Stock Market for every 1st October of every year. Table 2 shows the number of cases in NASDAQ Stock Market for every 12th October of every year. Table 3 shows the number of cases in Standard & Poor's (S&P) Stock market for every 10th January of every year. Table 4 shows the number of cases in Standard & Poor's (S&P) Stock market for every 10th October of every year.

Table 1. The number of cases in NASDAQ Stock Market for every 1st October of every year

| Date | Open | High | Low | Close | Volume | Adj Close |
|---|---|---|---|---|---|---|
| 01-10-1990 | 349 | 354 | 346 | 354 | 124380000 | 354 |
| 01-10-1991 | 527 | 528 | 525 | 528 | 162680000 | 528 |
| 01-10-1992 | 581 | 582 | 577 | 578 | 185130000 | 578 |
| 01-10-1993 | 76 | 763 | 761 | 763 | 290020000 | 763 |
| 01-10-1996 | 122 | 1227 | 1214 | 1221 | 542530000 | 1221 |
| 01-10-1997 | 1690 | 1696 | 1680 | 1690 | 970680000 | 1690 |
| 01-10-1998 | 1663 | 1693 | 1606 | 1612 | 856960000 | 1612 |
| 01-10-1999 | 2729 | 2739 | 2698 | 2736 | 973610000 | 2736 |
| 01-10-2001 | 1491 | 1491 | 1458 | 1480 | 1505140000 | 1480 |
| 01-10-2002 | 1180 | 1214 | 1160 | 1213 | 1707860000 | 1213 |
| 01-10-2003 | 1797 | 1832 | 1796 | 1832 | 1821740000 | 1832 |
| 01-10-2004 | 1909 | 1942 | 1908 | 1942 | 1820300000 | 1942 |
| 01-10-2007 | 2704 | 2743 | 2704 | 2740 | 1914080000 | 2740 |
| 01-10-2008 | 2075 | 2083 | 2046 | 2069 | 1899330000 | 2069 |
| 01-10-2009 | 2111 | 2112 | 2057 | 2057 | 2708170000 | 2057 |
| 01-10-2010 | 2386 | 2389 | 2359 | 2370 | 1932650000 | 2370 |
| 01-10-2012 | 3130 | 3146 | 3103 | 3113 | 1758170000 | 3113 |
| 01-10-2013 | 3774 | 3817 | 3774 | 3817 | 1843320000 | 3817 |
| 01-10-2014 | 4486 | 4486 | 4409 | 4422 | 2312630000 | 4422 |
| 01-10-2015 | 4624 | 4628 | 4559 | 4627 | 2133990000 | 4627 |

Table 2. The number of cases in NASDAQ Stock Market for every 12th October of every year

| Date | Open | High | Low | Close | Volume | Adj Close |
|---|---|---|---|---|---|---|
| 12-10-1990 | 326 | 327 | 323 | 327 | 132330000 | 327 |
| 12-10-1992 | 571 | 573 | 570 | 573 | 127850000 | 573 |
| 12-10-1993 | 772 | 772 | 770 | 772 | 316310000 | 772 |
| 12-10-1994 | 765 | 767 | 764 | 767 | 331570000 | 767 |
| 12-10-1995 | 1003 | 1015 | 1003 | 1015 | 421760000 | 1015 |
| 12-10-1998 | 1528 | 1560 | 1492 | 1546 | 764820000 | 1546 |
| 12-10-1999 | 2921 | 2923 | 2869 | 2872 | 1004090000 | 2872 |
| 12-10-2000 | 3241 | 3249 | 3071 | 3074 | 2128660000 | 3074 |
| 12-10-2001 | 1690 | 1707 | 1651 | 1703 | 2185970000 | 1703 |
| 12-10-2004 | 1913 | 1929 | 1904 | 1925 | 1508390000 | 1925 |
| 12-10-2005 | 2055 | 2064 | 2032 | 2037 | 2014750000 | 2037 |
| 12-10-2006 | 2318 | 2346 | 2318 | 2346 | 2003960000 | 2346 |
| 12-10-2007 | 2779 | 2806 | 2778 | 2805 | 1957790000 | 2805 |
| 12-10-2009 | 2145 | 2155 | 2128 | 2139 | 1784280000 | 2139 |
| 12-10-2010 | 2397 | 2421 | 2379 | 2417 | 1960920000 | 2417 |
| 12-10-2011 | 2606 | 2629 | 2602 | 2604 | 1967190000 | 2604 |
| 12-10-2012 | 3049 | 3061 | 3039 | 3044 | 1524840000 | 3044 |
| 12-10-2015 | 4839 | 4846 | 4818 | 4838 | 1343820000 | 4838 |

Table 3. The number of cases in Standard & Poor's (S&P) Stock market for every 10th January of every year

| Date | Open | High | Low | Close | Volume | Adj Close |
|---|---|---|---|---|---|---|
| 10-01-1950 | 17 | 17 | 17 | 17 | 2160000 | 17.03 |
| 10-01-1951 | 20 | 20 | 20 | 20 | 3270000 | 20.85 |
| 10-01-1952 | 23 | 23 | 23 | 23 | 1520000 | 23.86 |
| 10-01-1955 | 35 | 35 | 35 | 35 | 4300000 | 35.79 |
| 10-01-1956 | 44 | 44 | 44 | 44 | 2640000 | 44.16 |

| 10-01-1957 | 46 | 46 | 46 | 46 | 2470000 | 46.27 |
| 10-01-1958 | 40 | 40 | 40 | 40 | 2010000 | 40.37 |
| 10-01-1961 | 58 | 58 | 58 | 58 | 4840000 | 58.97 |
| 10-01-1962 | 69 | 69 | 68 | 68 | 3300000 | 68.96 |
| 10-01-1963 | 64 | 65 | 64 | 64 | 4520000 | 64.71 |
| 10-01-1964 | 76 | 76 | 75 | 76 | 5260000 | 76.24 |
| 10-01-1966 | 93 | 93 | 92 | 93 | 7720000 | 93.33 |
| 10-01-1967 | 82 | 83 | 82 | 82 | 8120000 | 82.81 |
| 10-01-1968 | 96 | 97 | 95 | 96 | 11670000 | 96.52 |
| 10-01-1969 | 101 | 102 | 100 | 100 | 12680000 | 100.93 |
| 10-01-1972 | 103 | 103 | 102 | 103 | 15320000 | 103.32 |
| 10-01-1973 | 119 | 120 | 118 | 119 | 20880000 | 119.43 |
| 10-01-1974 | 93 | 94 | 91 | 92 | 16120000 | 92.39 |
| 10-01-1975 | 71 | 73 | 71 | 72 | 25890000 | 72.61 |
| 10-01-1977 | 105 | 105 | 104 | 105 | 20860000 | 105.2 |
| 10-01-1978 | 90 | 91 | 89 | 90 | 25180000 | 90.17 |
| 10-01-1979 | 99 | 99 | 98 | 98 | 24990000 | 98.77 |
| 10-01-1980 | 109 | 110 | 108 | 109 | 55980000 | 109.89 |
| 10-01-1984 | 168 | 169 | 167 | 167 | 109570000 | 167.95 |
| 10-01-1985 | 165 | 168 | 164 | 168 | 124700000 | 168.31 |
| 10-01-1986 | 206 | 207 | 205 | 205 | 122800000 | 205.96 |
| 10-01-1989 | 280 | 281 | 279 | 280 | 140420000 | 280.38 |
| 10-01-1990 | 349 | 349 | 344 | 347 | 175990000 | 347.31 |
| 10-01-1991 | 311 | 314 | 311 | 314 | 124510000 | 314.53 |
| 10-01-1992 | 417 | 417 | 413 | 415 | 236130000 | 415.1 |
| 10-01-1994 | 469 | 475 | 469 | 475 | 319490000 | 475.27 |
| 10-01-1995 | 460 | 464 | 460 | 461 | 352450000 | 461.68 |
| 10-01-1996 | 609 | 609 | 597 | 598 | 496830000 | 598.48 |
| 10-01-1997 | 754 | 759 | 746 | 759 | 545850000 | 759.5 |
| 10-01-2000 | 1441 | 1464 | 1441 | 1456 | 1064800000 | 1457.6 |
| 10-01-2001 | 1300 | 1313 | 1287 | 1313 | 1296500000 | 1313.27 |
| 10-01-2002 | 1155 | 1159 | 1150 | 1156 | 1299000000 | 1156.55 |
| 10-01-2003 | 927 | 932 | 917 | 927 | 1485400000 | 927.57 |
| 10-01-2005 | 1186 | 1194 | 1184 | 1192 | 1490400000 | 1190.25 |
| 10-01-2006 | 1290 | 1290 | 1283 | 1289 | 2373080000 | 1289.69 |
| 10-01-2007 | 1408 | 1415 | 1405 | 1414 | 2764660000 | 1414.85 |
| 10-01-2008 | 1406 | 1429 | 1395 | 1420 | 5170490000 | 1420.33 |
| 10-01-2011 | 1270 | 1271 | 1262 | 1269 | 4036450000 | 1269.75 |
| 10-01-2012 | 1280 | 1296 | 1280 | 1292 | 4221960000 | 1292.08 |
| 10-01-2013 | 1461 | 1472 | 1461 | 1472 | 4081840000 | 1472.12 |
| 10-01-2014 | 1840 | 1843 | 1832 | 1842 | 3335710000 | 1842.37 |

Table 4. The number of cases in Standard & Poor's (S&P) Stock market for every 10th October of every year

Support vector machines are a group of algorithms that have been executed in classification, acknowledgment, regression and time series. Novel advanced-fuzzy support vector machine (NA-FSVM) began as a usage of Structural Risk Minimization standard to create parallel classifications. Novel advanced-

| Date | Open | High | Low | Close | Volume | Adj Close |
|---|---|---|---|---|---|---|
| 01-10-1951 | 23 | 23 | 23 | 23 | 1330000 | 23 |
| 01-10-1952 | 24 | 24 | 24 | 24 | 1060000 | 24 |
| 01-10-1953 | 23 | 23 | 23 | 23 | 940000 | 23 |
| 01-10-1954 | 32 | 32 | 32 | 32 | 1850000 | 32 |
| 01-10-1956 | 45 | 45 | 45 | 45 | 2600000 | 45 |
| 01-10-1957 | 43 | 43 | 43 | 43 | 1680000 | 43 |
| 01-10-1958 | 50 | 50 | 50 | 50 | 3780000 | 50 |
| 01-10-1959 | 57 | 57 | 57 | 57 | 2660000 | 57 |
| 01-10-1962 | 56 | 56 | 55 | 55 | 3090000 | 55 |
| 01-10-1963 | 72 | 73 | 72 | 72 | 4420000 | 72 |
| 01-10-1964 | 84 | 85 | 84 | 84 | 4470000 | 84 |
| 01-10-1965 | 90 | 90 | 89 | 90 | 7470000 | 90 |
| 01-10-1968 | 103 | 104 | 102 | 103 | 15560000 | 103 |
| 01-10-1969 | 93 | 94 | 92 | 93 | 9090000 | 93 |

| | | | | | | | | | | | | | |
|---|---|---|---|---|---|---|---|---|---|---|---|---|---|
| 01-10-1970 | 84 | 85 | 83 | 84 | 9700000 | 84 | 01-10-1992 | 418 | 419 | 415 | 416 | 204780000 | 416 |
| 01-10-1971 | 98 | 99 | 98 | 99 | 13400000 | 99 | 01-10-1993 | 459 | 461 | 458 | 461 | 256880000 | 461 |
| 01-10-1973 | 108 | 109 | 107 | 108 | 15830000 | 108 | 01-10-1996 | 687 | 690 | 684 | 689 | 421550000 | 689 |
| 01-10-1974 | 64 | 64 | 62 | 63 | 16890000 | 63 | 01-10-1997 | 947 | 957 | 947 | 955 | 598660000 | 955 |
| 01-10-1975 | 84 | 85 | 83 | 83 | 14070000 | 83 | 01-10-1998 | 1017 | 1017 | 981 | 986 | 899700000 | 986 |
| 01-10-1976 | 105 | 106 | 104 | 104 | 20620000 | 104 | 01-10-1999 | 1283 | 1283 | 1266 | 1283 | 896200000 | 1283 |
| 01-10-1979 | 109 | 109 | 108 | 109 | 24980000 | 109 | 01-10-2001 | 1041 | 1041 | 1027 | 1039 | 1175600000 | 1039 |
| 01-10-1980 | 125 | 128 | 125 | 127 | 48720000 | 127 | 01-10-2002 | 815 | 848 | 813 | 848 | 1780900000 | 848 |
| 01-10-1981 | 116 | 118 | 115 | 117 | 41600000 | 117 | 01-10-2003 | 996 | 1018 | 996 | 1018 | 1566300000 | 1018 |
| 01-10-1982 | 120 | 122 | 120 | 122 | 65000000 | 122 | 01-10-2004 | 1115 | 1132 | 1115 | 1132 | 1582200000 | 1132 |
| 01-10-1984 | 166 | 166 | 164 | 165 | 73630000 | 165 | 01-10-2007 | 1527 | 1549 | 1527 | 1547 | 3281990000 | 1547 |
| 01-10-1985 | 182 | 185 | 182 | 185 | 130200000 | 185 | 01-10-2008 | 1164 | 1167 | 1141 | 1161 | 5782130000 | 1161 |
| 01-10-1986 | 231 | 235 | 231 | 234 | 143600000 | 234 | 01-10-2009 | 1055 | 1055 | 1029 | 1030 | 5791450000 | 1030 |
| 01-10-1987 | 322 | 327 | 322 | 327 | 193200000 | 327 | 01-10-2010 | 1143 | 1150 | 1139 | 1146 | 4298910000 | 1146 |
| 01-10-1990 | 306 | 315 | 306 | 315 | 202210000 | 315 | 01-10-2012 | 1441 | 1457 | 1441 | 1444 | 3505080000 | 1444 |
| 01-10-1991 | 388 | 390 | 388 | 389 | 163570000 | 389 | 01-10-2013 | 1682 | 1697 | 1682 | 1695 | 3238690000 | 1695 |

fuzzy support vector machine (NA-FSVM) rose up out of research in statistical learning hypothesis on the best way to manage speculation, and locate an ideal exchange off between structural many-sided quality and observational risk. Novel advanced-fuzzy support vector machine (NA-FSVM) classify points by assigning them to one of two disjoint half spaces, either in the example space or in a higher-dimensional element space.

**Investigational outcomes**
The four parameter combinations and corresponding forecast accuracies are given in Table 1, 2, 3 and 4. Four parameter combinations given in each of the four tables are assumed to be the best ones in representing all cases in the whole information set. With these parameter combinations, authors are presently ready to perform comparison experiments of the Novel advanced-fuzzy support vector machine (NA-FSVM). The information sets summarized in the tables 3 were connected to the Novel advanced-fuzzy support vector machine (NA-FSVM) with four diverse parameter combinations. Combinations are around the same. Nonetheless, since its normal holdout execution is generally more prominent than the others, the execution of the third parameter is moderately superior to others. Along these lines, the forecast execution of this parameter mix can be received as the best of the Fuzzy support vector machine show. For the selected parameter mix, the best holdout execution (88) was gotten in 2013 while the worst one was acquired in 1957. The forecast execution of spiral basis Novel advanced-fuzzy support vector machine (NA-FSVM) model is diverse for four parameter combinations and is less than that of the Fuzzy support vector machine demonstrate. The results showed that the Novel advanced-fuzzy support vector machine (NA-FSVM) gives a superior forecast execution than the fuzzy support vector machine display.

Figure 2. (left) Graphical representation showing the cases in NASDAQ Stock Market for every 1st October of every year

Figure 3. (right) Graphical representation showing the cases in NASDAQ Stock Market for every 12th October of every year

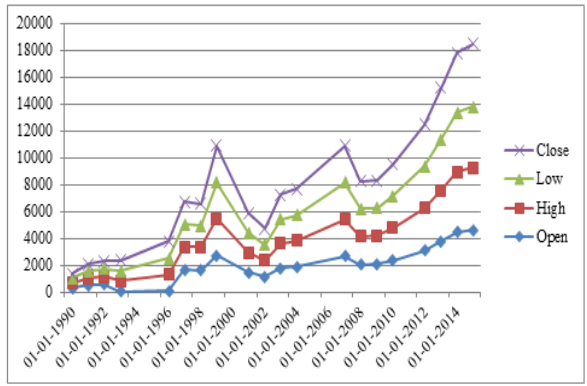 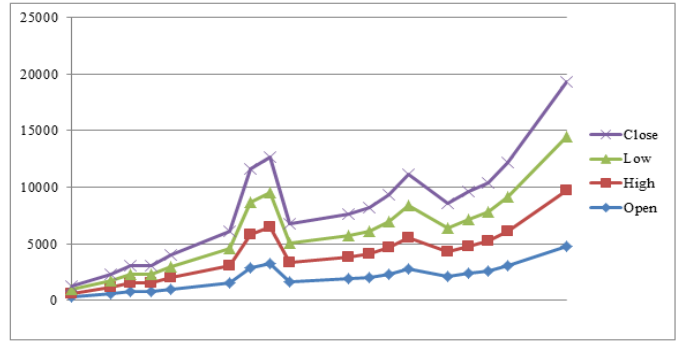

**Conclusion:**

Foreseeing the bearing of movements of the stock market list is essential for the advancement of powerful market exchanging strategies. It usually affects a money related broker's decision to purchase or sell an instrument. Successful forecast of stock prices may promise alluring benefits for investors. These tasks are profoundly entangled and exceptionally troublesome. This study endeavored to anticipate the bearing of stock value development in the NASDAQ Stock Market, Standard and Poor's (S&P) Stock market.

Two expectation models of NASDAQ Stock Market, Standard and Poor's (S&P) Stock market were constructed and their performances were thought about on the day by day information from 1950 to 2015. Based on the test results got, some vital conclusions can be drawn. First of all, it should be emphasized that both the Fuzzy-support vector machine and NA-FSVM models showed significant execution in foreseeing the heading of stock value development. Thus, we can say that both the NA-FSVM is useful forecast tools for this point. The normal forecast execution of the NA-FSVM demonstrates 82.3% was discovered significantly superior to anything that of the Fuzzy-support vector machine show 79. 3%. To the best information of the authors, the expectation execution of the proposed models outperforms similar studies in the writing. Notwithstanding, forecast performances of our models might be enhanced by two ways. The first is to adjust the model parameters by leading a more sensitive and comprehensive parameter setting, which can be a future work for interested

The principal feature is to regulate the classic parameters by leading a more sensitive and comprehensive parameter setting, which can be a future work for interested readers. Second, extraordinary or extra info variables can be used in the models. In spite of the fact that we received ten specialized indicators, some other large scale financial variables such as outside trade rates, interest rates and consumer value record and so on can be used as inputs of the models. Nevertheless, ten specialized indicators received here demonstrated that they are useful in anticipating the course of stock value development. Another imperative issue that should be specified here is the differences among the expectation performances for every year. Under such circumstances of crisis, a decrease in the expectation execution of specialized indicators can be considered worthy.

**References**

[1] Aydogdu, Mahmut, and Mahmut Firat. "Estimation of failure rate in water distribution network using fuzzy clustering and LS-SVM methods." Water Resources Management 29, no. 5 (2015): 1575-1590.

[2] Chen, Pu, and Dayong Zhang. "Constructing support vector machines ensemble classification method for imbalanced datasets based on fuzzy integral." In International Conference on Industrial, Engineering and Other Applications of Applied Intelligent Systems, pp. 70-76. Springer International Publishing, 2014.

[3] Cheng, Wei-Yuan, and Chia-Feng Juang. "A fuzzy model with online incremental SVM and margin-selective gradient descent learning for classification problems." IEEE Transactions on Fuzzy Systems 22, no. 2 (2014): 324-337.

[4] Esme, Engin, and Bekir Karlik. "Fuzzy c-means based support vector machines classifier for perfume recognition." Applied Soft Computing 46 (2016): 452-458.

[5] Grigoryan, Hakob. "A Stock Market Prediction Method Based on Support Vector Machines (SVM) and Independent Component Analysis (ICA)." Database Systems Journal 7, no. 1 (2016): 12-21.

[6] Hipni, Afiq, Ahmed El-shafie, Ali Najah, Othman Abdul Karim, Aini Hussain, and Muhammad Mukhlisin. "Daily forecasting of dam water levels: comparing a support vector machine (SVM) model with adaptive neuro fuzzy inference system (ANFIS)." Water resources management 27, no. 10 (2013): 3803-3823.

[7] Iraji, Mohammad Saber, Mohammad Bagher Iraji, Alireza Iraji, and Razieh Iraji. "Age Estimation Based on CLM, Tree Mixture With Adaptive Neuron Fuzzy, Fuzzy Svm." International Journal of Image, Graphics and Signal Processing 6, no. 3 (2014): 51.

[8] Melin, Patricia, and Oscar Castillo. "A review on type-2 fuzzy logic applications in clustering, classification and pattern recognition." Applied soft computing 21 (2014): 568-577.

[9] Moustakidis, Serafeim, Giorgos Mallinis, Nikos Koutsias, John B. Theocharis, and Vasilios Petridis. "SVM-based fuzzy decision trees for classification of high spatial resolution remote sensing images." IEEE Transactions on Geoscience and Remote Sensing 50, no. 1 (2012): 149-169.

[10] Nair, Binoy B. "Empirical investigations on soft computing based approaches to design of stock trading recommender systems." (2015).

[11] Wang, H., & Wang, J. (2014, November). An effective image representation method using kernel classification. In *2014 IEEE 26th International Conference on Tools with Artificial Intelligence* (pp. 853-858). IEEE.

[12] Palma, Giovanni, Isabelle Bloch, and Serge Muller. "Detection of masses and architectural distortions in digital breast tomosynthesis images using fuzzy and a contrario approaches." Pattern Recognition 47, no. 7 (2014): 2467-2480.

[13] Punithavathani, D. Shalini, and Sheryl Radley. "Performance analysis for wireless networks: An analytical approach by multifarious sym teredo." The Scientific World Journal 2014 (2014).

[14] Ramanathan, Thirumalaimuthu Thirumalaiappan, and Dharmendra Sharma. "An SVM-Fuzzy Expert System Design For Diabetes Risk Classification." International Journal of Computer Science and Information Technologies 6, no. 3 (2015): 2221-2226.

[15] Sriwastava, Brijesh Kumar, Subhadip Basu, and Ujjwal Maulik. "Protein–Protein interaction site prediction in Homo sapiens and E. coli using an interaction-affinity based membership function in fuzzy SVM." Journal of biosciences 40, no. 4 (2015): 809-818.

[16] Ticknor, J. L. (2013). A Bayesian regularized artificial neural network for stock market forecasting. Expert Systems with Applications, 40(14), 5501-5506.

[17] Yılmaz, Tayfur, and Bayram Kılıç. "Forecasting Bist 100 Index Using Artificial Neural Networks (ANN) And Adaptive Neuro Fuzzy Inference System (ANFIS) Method." Journal of Applied Research in Finance and Economics 2, no. 1 (2016): 18-27.